# STRUCTURING CAUSAL TREE MODELS WITH CONTINUOUS VARIABLES *

**Lei Xu**
Department of Automation
Tsinghua University
Beijing China

**Judea Pearl**
Cognitive Systems Laboratory
Computer Science Department
UCLA, Los Angeles, CA. 90024-1600

## ABSTRACT

This paper considers the problem of invoking auxiliary, unobservable variables to facilitate the structuring of causal tree models for a given set of continuous variables. Paralleling the treatment of bi-valued variables in [Pearl 1986], we show that if a collection of coupled variables are governed by a joint normal distribution and a tree-structured representation exists, then both the topology and all internal relationships of the tree can be uncovered by observing pairwise dependencies among the observed variables (i.e., the leaves of the tree). Furthermore, the conditions for normally distributed variables are less restrictive than those governing bi-valued variables. The result extends the applications of causal tree models which were found useful in evidential reasoning tasks.

## I. INTRODUCTION

Belief networks are directed acyclic graphs in which nodes represent propositional variables, the arcs signify direct dependencies between the linked propositions, and the strengths of these dependencies are quantified by conditional probabilities. Belief networks can be used to represent the generic knowledge of a domain expert, as well as inferece engines that manipulate this knowledge in evidential reasoning applications [Pearl 1986]. In particular, it was shown that in a singly-connected (e.g. tree-structured) network, beliefs can be updated coherently by local propagation through a network of parallel and autonomous processors, and that equilibrium is guaranteed to be reached in time proportion to the network diameter.

Since the efficacy of the scheme is based on a singly-connected (especially tree-structured) network, Pearl has proposed a preprocessing approach which introduces auxiliary variables and permanently turns multiply-connected belief networks into a tree. Naturally, the question arises as to whether it is possible to reconfigure every belief network as a tree. In [Pearl, 1986] it is shown that if all variables are bi-valued and if there exists a decomposition into a tree-structured network with auxiliary variables, then the topology of the tree can be uncovered uniquely from the observed correlations between pairs of variables. However, if the variables are not bi-valued, the problem is more complicated and remains an open question.

In this paper, the problems of structuring causal trees with continuous variables are considered. We show that if all the variables are normally distributed and if the activities of the visible variables are governed by a tree-decomposable joint normal distribution, then the tree can be structured from the observed correlations between pairs of variables. Moreover, the con-ditions for normally distributed variables to be tree-decomposable are less restrictive than the corresponding conditions for bi-valued variables.

---

* This work was supported in part by the National Science Foundation Grant, DCR 83-13875.

## II. THEORETICAL BACKGROUND

### A. Nomenclature and Problem Statement

Let $x_1, x_2, ..., x_n$ be random variables from a $n$-dimensional joint normal distribution:

$$f(x_1, x_2, ..., x_n) = (2\pi)^{-\frac{n}{2}} (det \Sigma_n)^{-\frac{1}{2}} \exp[-\tfrac{1}{2}(\mathbf{x}_n - \mu_n)^t \Sigma_n^{-1}(\mathbf{x}_n - \mu_n)] \quad (1)$$

Where $\mathbf{x}_n = (x_1, x_2, ..., x_n)^t$, $\mu_n = E\mathbf{x}_n$ is the mean vector and $\Sigma_n = E(\mathbf{x}_n - \mu_n)(\mathbf{x}_n - \mu_n)^t$ is the covariance matrix of $\mathbf{x}_n$.

Analogous with Section 3.2 in [Pearl, 1986], we can ask if $f(x_1, x_2, ..., x_n)$ can be represented as a marginal of an $n+1$ dimensional normal distribution of variables $\mathbf{x}_{n+1} = (w, \mathbf{x}_n^t)^t$ such that the $x_i$'s are conditionally independent given $w$, i.e.

$$f(x_1, x_2, ..., x_n) = \int_{-\infty}^{+\infty} f_s(x_1, x_2, ..., x_n, w) \, dw \quad (2)$$

$$f_s(x_1, x_2, ..., x_n, w) = \prod_{i=1}^{n} f_s(x_i | w) \, f(w) \quad (3)$$

Where $f_s(x_i | w)$, $i = 1, ..., n$ relate each $x_i$ to the central hidden variable $w$ (see Figure 1).

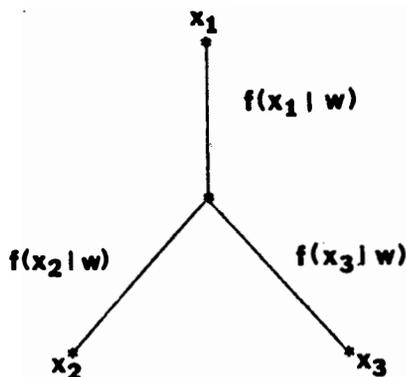

**Figure 1**

If the decomposition in (2) is possible, we name $f_s$ a *star-distribution* and call $f$ *star-decomposable*.

Instead of one hidden variable, $w$, we can use $m$ hidden variables ($m \leq n - 2$) to form a tree-like structure (see Figure 2), in which each triplet of leaves forms a star, but the central variable may differ from triplet to triplet.

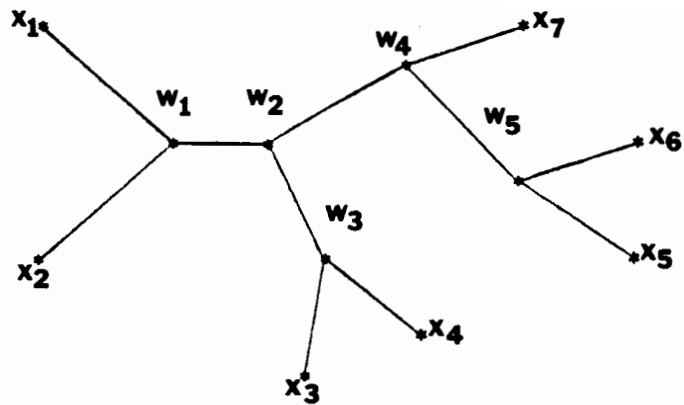

**Figure 2**

A normal distribution $f(x_1, x_2, ..., x_n)$ is said to be *tree-decomposable* if it is a marginal of an $n + m$ dimensional normal distribution:

$$f_t(x_1, x_2, ..., x_n, w_1, w_2, ..., w_m) \quad m \leq n-2 \tag{4}$$

that maps into a tree; i.e., $w_1, w_2, ..., w_m$ correspond to the internal nodes of a tree and $x_1, x_2, ..., x_n$ to its leaves.

In this paper we assume that a) each $w$ has at least three neighbors and b) $f(x|y) \neq f(x)$ for each $x, y \in \{x_1, x_2, ..., x_n, w_1, w_2, ..., w_m\}$, i.e., there are genuine dependencies between the linked variables (otherwise the tree can be decomposed into a forest).

The problem is whether $f_t(x_1, x_2, ..., x_n, w_1, w_2, ..., w_m)$ can be recovered given a tree-decomposable normal distribution $f(x_1, x_2, ..., x_n)$.

## B. A Theorem on Normal Distributions

It is well known (e.g., [Gigi 1977]) that the covariance matrix $\Sigma$ of a normal random vector $\mathbf{x} = (x_1, x_2, ..., x_p)^t$ is a $p \times p$ diagonal matrix, iff the components of $\mathbf{x}$ are independent normal random variables. Moreover,

**Theorem 1:** Let $\mathbf{x} = (\mathbf{x}_{(1)}^t, \mathbf{x}_{(2)}^t)^t$, $\mathbf{x}_{(1)}^t = (x_1 \cdots x_q)$, $\mathbf{x}_{(2)}^t = (x_{q+1} \cdots x_p)$
Let $\mu = E\mathbf{x}$ be similarly partitioned as $\mu = (\mu_{(1)}, \mu_{(2)})^t$ and let $\Sigma$ be partitioned as

$$\Sigma = \begin{Bmatrix} \Sigma_{11} & \Sigma_{12} \\ \Sigma_{21} & \Sigma_{22} \end{Bmatrix} \tag{5}$$

where $\Sigma_{11}$ is the $q \times q$ upper left-hand corner submatrix of $\Sigma$. If $\mathbf{x}$ is normally distributed with mean $\mu$ and covariance matrix $\Sigma$, then

a. The vectors $\mathbf{x}_{(1)}$ and $\mathbf{x}_{(2)} - \Sigma_{21} \Sigma_{11}^{-1} \mathbf{x}_{(1)}$ are independently normally distributed with means $\mu_{(1)}, \mu_{(2)} - \Sigma_{21} \Sigma_{11}^{-1} \mu_{(1)}$, and covariance matrices

$\Sigma_{11}$, $\Sigma_{22 \cdot 1} = \Sigma_{22} - \Sigma_{21} \Sigma_{11}^{-1} \Sigma_{12}$, respectively.

b. The marginal distribution of $x_{(1)}$ is q-variate normal with mean $\mu_{(1)}$ and covariance matrix $\Sigma_{11}$.

c. The conditional distribution of $x_{(2)}$ given $x_{(1)}$ is normal with mean $\mu_{(2)} + \Sigma_{21}\Sigma_{11}^{-1}(x_{(1)} - \mu_{(1)})$ and covariance matrix $\Sigma_{22 \cdot 1}$.

The proof of the theorem is given in [Gigi, 1977, pp. 51-53] and it will be used in the next section.

## III. STAR-DECOMPOSABLE TRIPLETS OF GAUSSIAN VARIABLES

Let $f(x_1, x_2, x_3)$ be a 3-dimensional joint normal distribution as in (1), for $n = 3$ with mean $\mu = (\mu_1 \, \mu_2 \, \mu_3)^t$ and covariance matrix

$$\Sigma = \begin{vmatrix} \sigma_{11} & \sigma_{12} & \sigma_{13} \\ \sigma_{21} & \sigma_{22} & \sigma_{23} \\ \sigma_{31} & \sigma_{32} & \sigma_{33} \end{vmatrix} \quad \text{and} \quad \sigma_{ij} = \sigma_{ji}, \, i, j = 1, 2, 3 \tag{6}$$

If $f(x_1, x_2, x_3)$ is star-decomposable, then it is a marginal of a 4-dimensional joint normal distribution $f_s(w, x_1, x_2, x_3)$ with mean $\mu_s = (\mu_w, \mu^t)^t$ and covariance matrix

$$\Sigma_s = \begin{array}{|c|ccc|} \hline \sigma_{ww} & \sigma_{1w} & \sigma_{2w} & \sigma_{3w} \\ \hline \sigma_{w1} & & & \\ \sigma_{w2} & & \Sigma & \\ \sigma_{w3} & & & \\ \hline \end{array}$$

and

$$\sigma_{wi} = \sigma_{iw}, \, i = 1, 2, 3$$

$$f(w, x_1, x_2, x_3) = f(x_1, x_2, x_3 | w) f(w) \tag{7}$$

$$f(x_1, x_2, x_3, | w) = f(x_1 | w) f(x_2 | w) f(x_3 | w) \tag{8}$$

Theorem 1 states that $f(x_1, x_2, x_3 | w)$, $f(w)$ and $f(x_i | w)$'s are also normal distributions, and the mean vector and covariance matrix of $f(x_1, x_2, x_3 | w)$ are given by

$$\mu_{1 \cdot 2 \cdot 3 | w} = \mu - \sigma_{ww}^{-1} \left[ \sigma_{w_1} \, \sigma_{w_2} \, \sigma_{w_3} \right]^t (w - \mu_w) \tag{9}$$

$$\Sigma_{1 \cdot 2 \cdot 3 | w} = \Sigma - \sigma_{ww}^{-1} \left[ \sigma_{w_1} \, \sigma_{w_2} \, \sigma_{w_3} \right]^t \left[ \sigma_{w_1} \, \sigma_{w_2} \, \sigma_{w_3} \right] \tag{10}$$

Additionally, the conditional independence stated in (8) implies that $\Sigma_{1\cdot 2\cdot 3|w}$ must be a diagonal matrix, thus

$$\sigma_{ij} - \sigma_{wi}\sigma_{jw}/\sigma_{ww} = 0, \quad i \neq j \text{ and } i,j = 1, 2, 3. \tag{11}$$

and

$$\sigma_{ii} - \sigma_{iw}^2/\sigma_{ww} > 0, \quad i = 1, 2, 3. \tag{12}$$

Using the correlation coefficients defined as

$$\rho_{ij} = \sigma_{ij}/(\sigma_{ii}\sigma_{ij})^{1/2} \tag{13}$$

(11) and (12) can be written as

$$\rho_{ij} = \rho_{iw}\rho_{jw}, \quad \text{for all } i,j \tag{14}$$

$$\rho_{iw}^2 \leq 1 \quad \text{for all } i. \tag{15}$$

Solving (14) for $\rho_{iw}$, we obtain

$$\rho_{1w} = (\rho_{12}\rho_{13}/\rho_{23})^{1/2} \quad \rho_{2w} = (\rho_{12}\rho_{23}/\rho_{13})^{1/2} \quad \rho_{3w} = (\rho_{13}\rho_{23}/\rho_{12})^{1/2} \tag{16}$$

The requirement that the $\rho_{iw}$'s must be real numbers with magnitude not exceeding 1, yields the following two conditions for $f(x_1, x_2, x_3)$ to be star-decomposable:

a.   $\rho_{12}, \rho_{13}, \rho_{23}$ are all positive, or two are negative and one is positive. In other words, the triplet $(x_1, x_2, x_3)$ is positively correlated.

b.   $\rho_{jk} \geq \rho_{ji}\rho_{ik}$ for all $i,j,k \in \{1,2,3\}$ and $i \neq j \neq k$

Summarizing the analysis above, we obtain Theorem 2.

**Theorem 2:**

1.   A necessary and sufficient condition for three random variables with a joint normal distribution to be star-decomposable is that the correlation coefficients satisfy the inequalities:

$$\rho_{jk} \geq \rho_{ji}\rho_{ik} \quad \rho_{ij}\rho_{ji}\rho_{ki} \geq 0 \tag{17}$$

for all $i,j,k \in \{1,2,3\}$, and $i \neq j \neq k$.

2.   $f(x_i|w) \sim N(\mu_{i|w}, \sigma_{i|w})$, $i = 1,2,3$ are specified by the parameters $\sigma_{i|w} = \sigma_{ii}(1 - \rho_{iw}^2) = \sigma_{ii}(1 - \rho_{ji}\rho_{ik}/\rho_{jk})$

$\mu_{i|w} = \mu_i - \sigma_{wi}(w - \mu_w)/\sigma_{ww} = \mu_i - \rho_{wi}\sqrt{\dfrac{\sigma_{ij}}{\sigma_{ww}}}(w - \mu_w)$ and

$f(w) \sim N(\mu_w, \sigma_w)$, where $\sigma_{ww} > 0$ and $\mu_w$ may be chosen arbitrarily.

Part 2 of Theorem 2 can be proved from (9), (10) and (11) combined with (13) and (16).

In a manner similar to [Pearl, 1986] we may pose the following problem. Suppose $f(x_1, x_2, x_3)$ is an arbitrary distribution (not necessarily normal), can it be approximated by a star-decomposable normal distribution $f_s(x_1, x_2, x_3)$ which will have the same covariance matrix as $f$. The answer is implied by Theorem 2, and stated in Theorem 3.

**Theorem 3:** A necessary and sufficient condition for the second order dependencies among the triplets $x_1, x_2, x_3$ to support a star-decomposable normal joint distribution is that all the correlation coefficients obey the triangle inequality:

$$\rho_{12}\rho_{13}\rho_{23} \geq 0 \text{ and } \rho_{jk} \geq \rho_{ji}\rho_{ik} \text{ for all } i,j,k \in \{1,2,3\}, i \neq j \neq k$$

**Discussion:**

1. Comparing Theorem 1 and 2 of [Pearl, 1986], we see that the conditions for a normally distributed triplet $x_1, x_2, x_3$ to be star-decomposable are less restrictive than those for dichotomous variables; there is no restriction corresponding to the 3rd order constraint imposed on $P_{ijk}$ in [Pearl, 1986].

2. Part 2 of Theorem 2 illustrates that if the conditions (17) are satisfied, the densities $f(w), f(x_i | w)$'s can be determined, depending on the selection of $\mu_w, \sigma_w$. In the case of star-decomposable triplets, one may simply specify $\sigma_w^2 = 1$ and $\mu_w = 0$ i.e., let the inner node have a standard normal distribution. For tree-decomposable structures with more than three variables, the selection of $\mu_w, \sigma_w$ for the intermediate variables should be made in a consistent manner, as shown in the next section.

3. To ensure that $n(>3)$ variables of a joint normal distribution be star-decomposable, (14) and (15) must lead to a consistent solution for all the $\rho_{iw}$'s.

4. A simple (degenerate) example of triplets which always satisfy (17) is given by 1st-order Markov variables $x_1, x_2, x_3$, governed by the covariance matrix

$$\Sigma = \begin{vmatrix} \sigma_{11} & \rho\sqrt{\sigma_{11}\sigma_{22}} & \rho^2\sqrt{\sigma_{11}\sigma_{33}} \\ \rho\sqrt{\sigma_{11}\sigma_{22}} & \sigma_{22} & \rho\sigma_{22}\sigma_{33} \\ \rho^2\sqrt{\sigma_{11}\sigma_{33}} & \rho\sqrt{\sigma_{22}\sigma_{33}} & \sigma_{33} \end{vmatrix} \quad 1 > \rho > 0$$

Since $\rho_{12} = \rho$, $\rho_{13} = \rho^2$ and $\rho_{32} = \rho$,

we have
$$\rho_{12} > \rho_{13}\rho_{32}$$
$$\rho_{23} > \rho_{12}\rho_{13}$$
$$\rho_{13} = \rho_{12}\rho_{23}$$

In this case the central variable $w$ coincides with $x_2$.

## IV. STRUCTURING CAUSAL TREES

Consider the 4 possible topologies of 4-tuple of leaves in a tree, as given in Figure 3.

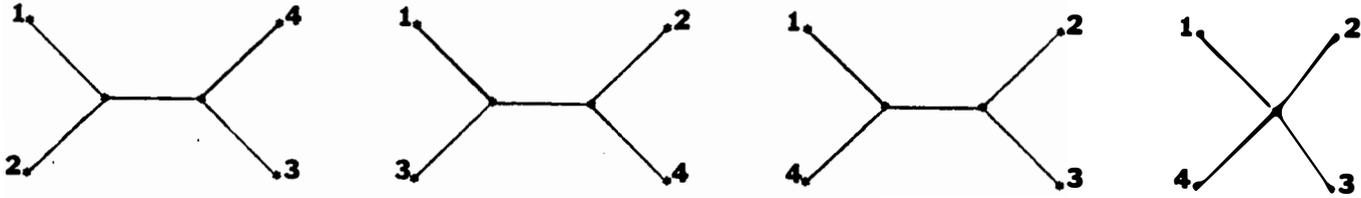

Figure 3

The topologies differ in the identity of the triplets which share a common center, e.g., in the topology of Figure 3 (a), [(1, 2, 3), (1, 2, 4)] share a common center; so does the pair [(1, 3, 4), (2, 3, 4)]. It follows that the star-decomposition of either triplet (1, 2, 3) or (1, 2, 4) should yield the same values of $\sigma_w, \mu_w$. This means that the correlation coefficients of the 4 variables with a topology of Figure 3 (a) should satisfy (16),

$$\rho_{1w}^2 = \rho_{12}\rho_{13} / \rho_{23} \quad \text{for} \quad (1, 2, 3)$$

$$\rho_{1w}^2 = \rho_{12}\rho_{14} / \rho_{24} \quad \text{for} \quad (1, 2, 4)$$

Thus,

$$\rho_{12}\rho_{13} / \rho_{23} = \rho_{12}\rho_{14} / \rho_{24}$$

or

$$\rho_{13}\rho_{24} = \rho_{14}\rho_{32} \tag{18}$$

(Enforcing the equality of $\rho_{2w}^2$ for (1, 2, 3) and (1, 2, 4), would yield the same equation). Similarly, from the pair [ (1, 3, 4), (2, 3, 4) ], we also obtain (18) (note $\rho_{ij} = \rho_{ji}$). Thus, the equality in (18) may be taken as the essential condition identifying the topology of Figure 3(a).

The equalities characterizing the 4 possible topologies of Figure 3 are given in table 1.

## Table 1. The Characteristic Equalities of the 4 Toplogies in Figure 3

| Fig. 3(a) | Fig. 3(b) | Fig. 3(c) | Fig. 3(d) |
|---|---|---|---|
| $\rho_{13}\rho_{24} = \rho_{14}\rho_{23}$ | $\rho_{12}\rho_{34} = \rho_{14}\rho_{23}$ | $\rho_{12}\rho_{34} = \rho_{13}\rho_{24}$ | two equalities holding simultaneously |

Table 1 may be taken as a tool for both testing if a 4-tuple is tree-decomposable and, if so, deciding its topology. Since the basic test to decide the toplogy of any 4-tuple is the same as that of binary variables, the topology of the entire tree of $n$ leaves can be decided by the procedure described in [Pearl, 1986].

The next stage is to determine all the $f(x_i | w)$ and $f(w_j | w_k)$ functions. The functions $f(x_i | w_j)$ assigned to the peripheral branches of the tree are determined directly from the star-decomposition of triplets involving adjacent leaves, and these can be used to determine the functions $f(w_j | w_k)$ connecting the hidden variables. In Figure 4, for example, the star-decomposition of $f(x_1, x_2, x_3)$ yields $f(x_1 | w_1), f(x_2 | w_1)$, while $f(x_1 | w_2)$ can be obtained from the star-decomposition of $(x_1, x_3, x_6)$. These are sufficient for determining $f(w_1 | w_2)$, via

$$f(x_1 | w_2) = \int_{w_1} f(x_1 | w_1) f(w_1 | w_2) dw_1 \qquad (19)$$

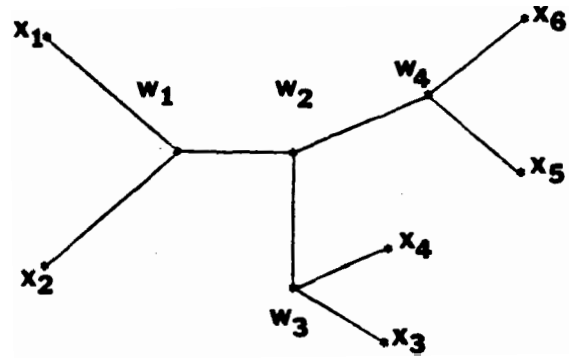

**Figure 4**

Let the mean and variance of $f(x_1 | w_2), f(x_1 | w_1)$ and $f(w_1 | w_2)$ be denoted by $\mu_{x_1 | w_2}, \sigma^2_{x_1 | w_2}, \mu_{x_1 | w_1}, \sigma^2_{x_1 | w_1}$ and $\mu_{w_1 | w_2}, \sigma^2_{w_1 | w_2}$ respectively, and let the mean vector and covariance matrix of $(x_1, w_1, w_2)$ be $[\mu_{x_1}, \mu_{w_1}, \mu_{w_2}]$ and

$$\begin{vmatrix} \sigma^2_{x_1} & \sigma_{x_1 w_1} & \sigma_{x_1 w_2} \\ \sigma_{w_1 x_1} & \sigma^2_{w_1} & \sigma_{w_1 w_2} \\ \sigma_{w_2 x_1} & \sigma_{w_1 w_2} & \sigma^2_{w_2} \end{vmatrix}$$

We will show that $\sigma_{w_1|w_2}$, $\mu_{w_1|w_2}$ (thus $f(w_1|w_2)$) may be determined from $f(x_1|w_2)$ and $f(x_1|w_1)$. From Theorem 1, the mean vector and covariance matrix of $f(x_1|w_1)f(w_1|w_2) = f(x_1, w_1|w_2)$ and $f(x_1, w_2 | w_1)$ are given by

$$\Sigma_{x_1 w_1 | w_2} = \begin{Bmatrix} \sigma_{x_1}^2 & \sigma_{x_1 w_1} \\ \sigma_{w_1 x_1} & \sigma_{w_1}^2 \end{Bmatrix} - \frac{1}{\sigma_{w_2}^2} \begin{Bmatrix} \sigma_{w_2 x_1} \\ \sigma_{w_2 w_1} \end{Bmatrix} \{\sigma_{w_2 x_1} \ \sigma_{w_2 w_1}\} \quad (20)$$

$$\Sigma_{x_1 w_2 | w_1} = \begin{Bmatrix} \sigma_{x_1}^2 & \sigma_{w_2 x_1} \\ \sigma_{x_1 w_2} & \sigma_{w_2}^2 \end{Bmatrix} - \frac{1}{\sigma_{w_1}^2} \begin{Bmatrix} \sigma_{w_1 x_1} \\ \sigma_{w_1 w_2} \end{Bmatrix} \{\sigma_{w_1 x_1} \ \sigma_{w_1 w_2}\}$$

$$\mu_{x_1 w_1 | w_2} = \begin{Bmatrix} \mu_{x_1} \\ \mu_{w_1} \end{Bmatrix} - \frac{1}{\sigma_{w_2}} \begin{Bmatrix} \sigma_{w_2 x_1} \\ \sigma_{w_2 w_1} \end{Bmatrix} (w_2 - \mu_{w_2}) \quad (21)$$

$$\mu_{x_1 w_2 | w_1} = \begin{Bmatrix} \mu_{x_1} \\ \mu_{w_2} \end{Bmatrix} - \frac{1}{\sigma_{w_1}} \begin{Bmatrix} \sigma_{w_1 x_1} \\ \sigma_{w_2 w_1} \end{Bmatrix} (w_1 - \mu_{w_1})$$

Therefore,

$$\sigma_{x_1 | w_2}^2 = \sigma_{x_1}^2 - \sigma_{w_2 x_1}^2 / \sigma_{w_2}^2 \quad (22)$$

$$\sigma_{w_1 | w_2}^2 = \sigma_{w_1}^2 - \sigma_{w_2 w_1}^2 / \sigma_{w_2}^2$$

$$\sigma_{x_1 | w_1}^2 = \sigma_{x_1}^2 - \sigma_{w_1 x_1}^2 / \sigma_{w_1}^2$$

$$\mu_{x_1 | w_2} = \mu_{x_1} - \frac{\sigma_{w_2 x_1}}{\sigma_{w_2}} (w_2 - \mu_{w_2}) \quad (23)$$

$$\mu_{w_1 | w_2} = \mu_{w_1} - \frac{\sigma_{w_2 w_1}}{\sigma_{w_2}} (w_2 - \mu_{w_2})$$

$$\mu_{x_1 | w_1} = \mu_{x_1} - \frac{\sigma_{w_1 x_1}}{\sigma_{w_1}} (w_1 - \mu_{w_1})$$

It is to be noted that $f(x_1 w_2 | w_1) = f(x_1 | w_1) f(x_2 | w_1)$, hence, $\Sigma_{x_1 w_2 | w_1}$ is a diagonal matrix which leads to

$$\sigma_{w_2 x_1} = \sigma_{w_1 x_1} \sigma_{w_2 w_1} / \sigma_{w_1}^2 \qquad (24)$$

As mentioned in Theorem 2, the mean and variance of the first inner variable may be assigned arbitrarily. We set $\sigma_{w_1}^2 = 1$ and $\mu_{w_1} = 0$, then join (22) and (24) and keep in mind that $\sigma_{x_1 | w_1}^2, \sigma_{x_1}^2, \sigma_{x_1 | w_2}^2$ and $\mu_{x_1 | w_2}, \mu_{x_1}, \mu_{x_2 | w_1}$ are known from $f(x_1 | w_1), f(x_1 | w_2)$ and $f(x_1)$, and $\sigma_{w_1 | w_2}^2, \sigma_{w_2}^2, \sigma_{w_1 w_2}, \sigma_{w_1 x_1}$ may be solved. In turn, $\mu_{w_2}, \mu_{w_1 | w_2}$ may be solved by (23). Consequently, $f(w_1 | w_2)$ is determined by $\sigma_{w_1 | w_2}, \mu_{w_1 | w_2}$; $f(w_2)$ is determined by $\sigma_{w_2}^2, \mu_{w_2}$. In the same manner, $f(w_3 | w_2)$, and $f(w_4 | w_2)$ may also be determined.

## V. CONCLUSIONS

The paper extends the auxiliary-variable method of constructing causal tree representations to continuous variables. It shows that if the visible variables are governed by a tree-decomposable joint normal distribution, then the tree can be structured from the observed correlations between pairs of variables. Furthermore, the conditions for tree-decomposable normal distribution are less restrictive than those of bi-valued variables. The results should extend the applicability of causal tree models to evidential reasoning tasks involving continuous signals and can also be used to discover the underlying causal structure behind ill understood phenomena.

## REFERENCES

[Pearl, 1986] Judea Pearl. *Fusion, Propagation and Structuring in Belief Networks,* "Artificial Intelligence," Vol. 29, No. 3, September 1986, pp. 241-288.

[Gigi, 1977] N.C. Gigi. *Multivariate Statistical Inference,* Academic Press, New York, 1977.